\documentclass[english,11pt,a4paper]{article}
\usepackage[T1]{fontenc}
\usepackage{graphicx}
\usepackage[margin=1in]{geometry}
\usepackage{amsmath, amssymb, amsfonts}
\usepackage{booktabs}
\usepackage{multirow}
\usepackage{natbib}
\usepackage{setspace}
\usepackage{hyperref}
\usepackage{caption}
\usepackage{subcaption}
\usepackage{ltxcmds}
\usepackage{tikz}
\usepackage{pgfplots}
\pgfplotsset{compat=1.18}
\usepgfplotslibrary{groupplots}

\begin{document}
	\begin{titlepage}
		\raggedright
		
		{\LARGE\bfseries
			Incorporating data drift to perform survival analysis on credit risk
			\par}
		
		\vspace{1.5em}
		
		{\normalsize
			Jianwei Peng (corresponding author) \\ ORCiD ID: 0009-0001-4383-8263 \\
			Humboldt-Universität zu Berlin, School of Business and Economics, Spandauer Str. 1, 10178 Berlin \\
			\texttt{E-mail: jianwei.peng@student.hu-berlin.de}
			\par}
		
		\vspace{1em}
		
		{\normalsize
			Stefan Lessmann \\
			Humboldt-Universität zu Berlin, School of Business and Economics, Spandauer Str. 1, 10178 Berlin \\
			Bucharest University of Economic Studies, 6 Piata Romana, 1st district Bucharest, 010374, Romania \\
			\texttt{E-mail: stefan.lessmann@hu-berlin.de}
			\par}
		
		\vfill
		
		{\normalsize
			Word count: 9915
		}
		
		\date{}
		
	\end{titlepage}

\begin{abstract}
	Survival analysis has become a standard approach for modelling time to default by time-varying covariates in credit risk. Unlike most existing methods that implicitly assume a stationary data-generating process, in practise, mortgage portfolios are exposed to various forms of data drift caused by changing borrower behaviour, macroeconomic conditions, policy regimes and so on. This study investigates the impact of data drift on survival-based credit risk models and proposes a dynamic joint modelling framework to improve robustness under non-stationary environments. The proposed model integrates a longitudinal behavioural marker derived from balance dynamics with a discrete-time hazard formulation, combined with landmark one-hot encoding and isotonic calibration. Three types of data drift (sudden, incremental and recurring) are simulated and analysed on mortgage loan datasets from Freddie Mac. The experimental evidence shows that the proposed landmark-based joint model consistently outperforms classical survival models, tree-based drift-adaptive learners and gradient boosting methods in terms of discrimination and calibration across all drift scenarios, which confirms the superiority of our model design.
\end{abstract}

\section*{Practitioner Summary}
Credit risk models help lenders monitor default risk, set lending policies and manage portfolio losses. However, models built on past data can become less reliable when borrower behaviour, interest rates, housing markets or policy conditions change. This study proposes a dynamic credit risk modelling framework that is designed for such unstable environments. The framework combines standard loan information with behavioural markers that compare a borrower’s actual repayment path with the scheduled amortisation path. The introduction of time dummies also reflects that loans observed at different stages or under different market conditions may have different baseline risks. For practitioners, the approach can be applied to routinely collected servicing data. In this case, regular updating of default probabilities as new monthly information becomes available, making it valuable for portfolio monitoring, early warning systems and stress testing. The behavioural marker improves forward portability by capturing repayment behaviour in an economically interpretable way rather than relying only on fixed historical patterns. At the same time, time-specific adjustments and probability calibration imply that the model should be recalibrated regularly, especially after major economic, regulatory or portfolio-composition changes. Overall, this study offers a practical way to allow survival-based credit risk models to be more adaptive, transparent and reliable under data drift.

\noindent\textbf{Keywords:} 
Survival Analysis; Data Drift; Credit Risk; Joint Models

\newpage
\section{Introduction}
Credit risk modelling plays a central role in retail banking, regulatory capital calculation and portfolio risk management. Especially, mortgage default prediction has attracted sustained attention due to the long maturities, high exposures and strong sensitivity to economic conditions inherent in housing finance. Traditional credit scoring approaches typically rely on static classification models that predict default within a fixed horizon based on borrower and loan characteristics observed at origination \citep{abi2025machine}. However, such approaches will gradually lose their effectiveness in managing mortgage portfolios if borrower behaviour evolves over time and default is viewed as a time-to-event outcome rather than a one-off binary state. As a result, survival analysis has become a standard framework for modelling time to default in credit risk, offering a natural way to censor varying loan lifetimes and dynamic prediction horizons \citep{stepanova2002survival,steyerberg2013prognosis}.

Currently, common survival-based credit risk models are applied in offline settings, requiring the whole training set to be present in memory before the start of training, while most of them are exposed to non-stationary data streaming environments in practise \citep{suarez2023survey}. Firstly, borrower repayment behaviour evolves over the loan lifecycle, reflecting income changes, refinancing incentives and behavioural fatigue. Secondly, portfolio composition changes over time due to survivorship effects, prepayments and selective attrition, leading to systematic differences between early-life and late-life loans. Thirdly, external conditions such as interest rate regimes, housing market cycles and macroeconomic shocks introduce concept drifts that alter both feature distributions and the relationship between predictors and default risk \citep{arellano2008default}. Empirical studies have shown that ignoring such drift can lead to severe performance degradation and miscalibration in credit risk models \citep{klinkenberg2004learning,gama2014survey}.

Recent work has explored various strategies to address non-stationarity in credit risk. Several studies extend survival models by incorporating time-varying covariates or piecewise hazards \citep{bellotti2013forecasting,dirick2017time}. Others adopt machine learning algorithms, such as gradient boosting or online tree ensembles, that are more flexible in capturing complex patterns and concept drift \citep{tarazodar2024mitigating}. While these methods offer improvements in certain settings, they also exhibit important limitations. \citet{sa2007survival} believe that classical survival models rely on strong assumptions such as proportional hazards and often struggle with data drift. Machine learning models, although powerful in static environments, typically lack an explicit time-to-event structure and produce poorly calibrated probabilities, particularly in highly imbalanced and drifting data streams. Moreover, most existing approaches treat longitudinal borrower information in an ad hoc manner, failing to exploit the underlying temporal structure of repayment behaviour \citep{philip2018improved}.

This study argues that robust default prediction under data drift requires an integrated modelling strategy that simultaneously accounts for longitudinal borrower behaviour, time-to-event dynamics and temporal heterogeneity in the risk distribution. To this end, we propose a dynamic joint modelling framework that combines a longitudinal behavioural component with a discrete-time survival model using a landmarking strategy. The longitudinal component summarises the evolution of repayment behaviour through a balance-based deviation marker, capturing both the level and trend of borrower performance over time. The survival component models the probability of default within a future horizon conditional on information available at a given landmark month, allowing for dynamic updating as new information becomes available. To further address non-stationarity, the model incorporates landmark-specific baseline adjustments via one-hot encoding and applies isotonic regression to recalibrate predicted probabilities under drift.

This study makes three main contributions to the literature on credit risk modelling. Firstly, to the best of our knowledge, this is the first study to systematically investigate the performance of survival-based credit risk models under multiple concept drift scenarios including sudden, incremental and recurring drift together with label drift, using a unified experimental framework. Secondly, we propose a landmark-based dynamic joint modelling approach that integrates a longitudinal behavioural marker derived from balance dynamics with a discrete-time hazard formulation, allowing default risk to be updated dynamically as new information becomes available. Thirdly, a joint landmark one-hot and isotonic calibration (LMISO) strategy is introduced to explicitly address temporal heterogeneity and probability miscalibration under data drift. Its effectiveness is empirically proved to be superior to classical survival models, machine learning methods and drift-adaptive online learners.

By bridging survival analysis, longitudinal modelling and data drift adaptation, this work contributes to the growing literature on dynamic credit risk modelling and offers practical insights for risk managers operating in non-stationary environments. Beyond mortgage default prediction, the proposed framework is applicable to a wide range of operational risk and reliability problems characterised by evolving behaviour and drifting data distributions.

The remainder of this paper is organised as follows. Section 2 reviews related work on survival analysis in credit risk, joint modelling and data drift handling. Section 3 introduces the proposed dynamic joint modelling framework, including the longitudinal design, landmark-based hazard formulation and calibration strategy. Section 4 describes the data from Freddie Mac, preprocessing steps, drift simulation and experimental settings. Section 5 presents the empirical results and comparative analysis. Finally, Section 6 concludes the paper and outlines directions for future research.

\section{Related Work}
Survival analysis has a long tradition in credit risk because it directly models time to default while accommodating censoring and varying loan lifetimes. In the past decade, large-scale empirical studies have reinforced survival models as a state-of-the-art complement to standard binary scoring, and have compared Cox-type models, accelerated failure time models and flexible hazard specifications under credit settings \citep{dirick2015akaike,dirick2017time}. Parallel strands of work have extended survival analysis for credit risk by incorporating more flexible hazard shapes and time-varying effects, motivated by lifecycle effects and shifting default environments \citep{dirick2019macro,djeundje2019dynamic}. At the same time, discrete-time hazard formulations have gained attention because they align naturally with monthly reporting cycles, support multi-period default forecasting and offer implementation stability via generalized linear modelling and modern regularization \citep{breeden2022multihorizon,noorian2015risk}. These developments are closely related to regulatory and accounting contexts such as IFRS 9, where estimating the term structure of default risk is central to expected credit loss calculations \citep{botha2025approaches}.

A key limitation of much survival-based credit risk modelling is the treatment of borrower behaviour as exogenous, static, or only weakly dynamic. In domains where repeated measurements are available, joint models for longitudinal and time-to-event data provide a principled framework to address endogeneity and measurement error by linking an event process to an underlying latent longitudinal process \citep{rizopoulos2014combining,andrinopoulou2021reflection}. Recent methodological work continues to refine joint modelling for dynamic prediction, including strategies for combining multiple longitudinal markers and improving predictive performance under complex marker dynamics \citep{hashemi2025dynamic}. In credit risk specifically, joint modelling has been increasingly explored as a way to integrate behavioural trajectories with default timing; for example, joint frameworks for longitudinal credit processes and discrete survival outcomes have been proposed and empirically assessed by \citet{medina2023joint}. Complementary to joint modelling, the landmarking paradigm has become a dominant approach for dynamic survival prediction: it constructs a sequence of prediction problems at pre-specified landmark times using information available up to each landmark. Recent comparative work argues that joint modelling and landmarking are the two principal strategies for dynamic survival prediction with longitudinal information, and systematically investigates their trade-offs \citep{signorelli2025pencal}.

While these survival and dynamic prediction methods address temporal structure, an orthogonal challenge in deployed credit risk systems is data drift (covariate, concept and label drift). In data streams and real-world deployments, drift refers to temporal changes in feature/label distributions or the conditional relationship between predictors and outcomes, which can severely degrade discrimination and calibration \citep{gama2014survey}. Numerous literature has formalized drift types (sudden, gradual, incremental, recurring), drift detection and supervision in streaming settings. Surveys emphasize that recurring drift and regime-switching remain difficult because models must not only adapt but also reuse previously learned concepts \citep{suarez2023survey}. More recent work highlights the importance of unsupervised and weakly supervised drift monitoring, where labels are delayed or sparse—a setting that resembles many credit risk monitoring problems \citep{hinder2024one}. In operational and business analytics contexts, drift detection and governance have also been framed as risk-tiered decision processes linking drift signals to intervention policies \citep{muhammad2025cortex}.

Regarding credit risk, drift has been recognized as a practical source of model instability, especially under macroeconomic regime changes. Studies have investigated incorporating time-varying macroeconomic variables to explain and reduce drift in credit scoring, showing that evolving macro conditions can be a major driver of concept drift and performance decay \citep{bellotti2009credit}. More directly, drift adaptation methods have been applied to credit scoring problems, including approaches based on local regions, reweighting and adaptive updating strategies \citep{nikolaidis2022credit}. These studies align with a broader discussion in credit analytics that machine learning models can be accurate in stationary settings but require explicit mechanisms such as monitoring, adaptation and calibration to remain reliable under data drift \citep{patchipala2023tackling}.

A widely used family of drift-mitigation methods in streaming machine learning is drift-adaptive decision trees and ensembles. \citet{gomes2017adaptive} design Adaptive Random Forests (ARF) that combine online bagging with per-tree drift detectors to trigger background learners and model replacement, achieving strong empirical performance across diverse drifting streams. Adaptive Hoeffding-tree variants and regularization strategies have also been proposed to enhance robustness and prevent overfitting or instability in evolving streams \citep{barddal2019learning}. These methods are attractive as benchmarks for drift adaptation because they are explicitly designed for non-stationary streaming data.

In addition to adapting predictors, drift and temporal change can distort predicted probabilities, making probability calibration a critical issue for risk management. Post-hoc calibration methods such as isotonic regression are widely used to correct miscalibration in probabilistic classifiers. Related work from \citet{ojeda2023calibrating} has emphasized their importance for reliable decision-making, particularly in imbalanced settings where probability accuracy, not just ranking, drives expected loss and capital decisions. Calibration facing non-stationary data streams has also become an active topic with empirical studies showing that real-world drift can strongly degrade calibration even when discrimination remains acceptable \citep{roschewitz2025we}. In the context of predictive modelling workflows, isotonic regression is often highlighted as a flexible non-parametric calibrator that significantly decreases Brier score without changing rank ordering, which is especially appealing when drift shifts the base rate of the event \citep{ojeda2023calibrating}.

Overall, existing literature motivates an integrated view: (i) default is naturally a time-to-event outcome with lifecycle effects, favouring survival and discrete-time hazard frameworks \citep{dirick2017time,botha2025approaches}; (ii) borrower performance contains longitudinal signals that can be endogenously linked to default timing via joint modelling or dynamically via landmarking \citep{tanner2021dynamic,medina2023joint}; (iii) model reliability in practice requires explicit treatment of drift and calibration, for which drift-adaptive stream learners and post-hoc calibrators are prominent candidates \citep{gomes2017adaptive,suarez2023survey,ojeda2023calibrating}. This study builds on these strands by evaluating survival prediction under multiple drift regimes and by proposing a landmark-based joint approach with explicit temporal adjustment and calibration mechanisms to improve robustness in drifting credit risk environments.

\section{Methodology}

\subsection{Longitudinal Marker}

\subsubsection{Behavioural marker: scheduled vs actual repayment}
Longitudinal data refer to data collected repeatedly from the same subjects over multiple time points, enabling the analysis of within-subject change and temporal dynamics. The appropriate longitudinal marker should be selected by evaluating predictors based on theoretical relevance, temporal stability, measurement reliability, sensitivity to change, and empirical evidence of predictive validity for the outcome of interest \citep{hedeker2006longitudinal}. The combination of loan balance and interest rate is considered by \citet{medina2023joint} to be excellent longitudinal data in the study of Freddie Mac mortgage loan datasets. The monthly servicing data provide the observed outstanding balance $UPB_{cur}(t)$ at month $t$, where $t$ denotes the loan age in months ($t=LoanAge$). To separate systematic amortisation from behavioural deviations, we decompose each loan’s balance into (i) a scheduled path implied by amortisation mechanics and (ii) an observed deviation from this path.

Let $r$ denote the monthly interest rate, $N$ the loan term in months, and $OrigUPB$ the original unpaid balance. Under standard amortisation with fixed rate and scheduled monthly payments, the scheduled outstanding balance at month $t$ is computed as:
\begin{equation}
	B_{sch}(t)=\frac{(1+r)^N-(1+r)^t}{(1+r)^N-1}\cdot OrigUPB.
	\tag{1}
\end{equation}
Equation (1) follows from the annuity identity: the remaining balance equals the present value of the remaining $N-t$ payments, which yields the closed form ratio of geometric series. Using $B_{sch}(t)$, we define the behavioural balance deviation percentage:
\begin{equation}
	BD\_pct(t)=\frac{UPB_{cur}(t)-B_{sch}(t)}{B_{sch}(t)}.
	\tag{2}
\end{equation}
The marker $BD\_pct(t)$ measures whether the borrower is behind ($BD\_pct(t)>0$) or ahead ($BD\_pct(t)<0$) relative to the scheduled amortisation path. This construction retains an economic interpretation and mitigates scale effects across loans with different $OrigUPB$ and terms.

\subsubsection{Loan-specific longitudinal trajectory (per-loan regression)}
To summarise longitudinal repayment behaviour with a low-dimensional representation, we fit a lightweight per-loan trajectory model using normalised time. For loan $i$, let $N_i$ be its term and define the normalised age $t/N_i\in[0,1]$. Denote the observed marker at month $t$ as $m_i(t)\equiv BD\_pct_i(t)$ and we obtain a linear trajectory as a computationally lighter random-effects proxy:
\begin{equation}
	m_i(t)\approx b_{0i}+b_{1i}\frac{t}{N_i}+\varepsilon_{it}.
	\tag{3}
\end{equation}
$\varepsilon_{it}$ is a residual term capturing month-to-month idiosyncratic fluctuations in $BD\_pct(t)$ that are not explained by the linear trajectory in normalised age. Under the standard regression assumption, $\mathbb{E}[\varepsilon_{it}\mid x_{it}]=0$, meaning the linear component explains the systematic part of the marker and the remaining noise is centred around zero. In this setting, small positive and negative deviations arise from reporting noise, minor payment timing variation and rounding, so $\varepsilon_{it}$ is expected to fluctuate around zero rather than exhibit persistent bias.

The coefficients $b_{0i}$ and $b_{1i}$ capture two critically meaningful behavioural dimensions: a level (early deviation) and a trend (deterioration or improvement as the loan seasons). Estimation is performed by closed-form OLS per loan using only observed months, with a small ridge term to stabilise estimates when the number of observations is small or the within-loan variation in $t/N_i$ is limited.

Let $x_{it}=t/N_i$ and $y_{it}=m_i(t)$. Collect the $n_i$ observed months for loan $i$. Define
\begin{equation}
	\bar{x}_i=\frac{1}{n_i}\sum_{t}x_{it},\qquad \bar{y}_i=\frac{1}{n_i}\sum_{t}y_{it},
	\tag{4}
\end{equation}
\begin{equation}
	S_{xx,i}=\sum_t (x_{it}-\bar{x}_i)^2,\qquad S_{xy,i}=\sum_t (x_{it}-\bar{x}_i)(y_{it}-\bar{y}_i).
	\tag{5}
\end{equation}
With a tiny ridge $\lambda>0$, the slope and intercept estimators are:
\begin{equation}
	\hat{b}_{1i}=\frac{S_{xy,i}}{S_{xx,i}+\lambda},\qquad 
	\hat{b}_{0i}=\bar{y}_i-\hat{b}_{1i}\bar{x}_i.
	\tag{6}
\end{equation}
The fitted longitudinal summary can be evaluated at any month $t$:
\begin{equation}
	\hat{m}_i(t)=\hat{b}_{0i}+\hat{b}_{1i}\frac{t}{N_i}.
	\tag{7}
\end{equation}
This $\hat{m}_i(t)$ plays the role of a posterior-like behavioural state, but is deliberately kept computationally light to support large-scale mortgage portfolios.

\subsection{Landmarking}
To enable dynamic prediction that updates as information accrues, we adopt a landmarking strategy as shown in Figure~\ref{fig:landmark_sampling_scheme}. Fix a landmark month $L$ and a prediction horizon $H$. For each loan $i$ that is still at risk at age $L$, we construct:
\begin{itemize}
	\item observed covariates $X_i(L)$ (static + time-varying observed at $L$),
	\item longitudinal summary $\hat{m}_i(L)$,
	\item outcome: future default within horizon $H$.
\end{itemize}
The target is:
\begin{equation}
	Y^{(L,H)}_i=
	\begin{cases}
		1, & \text{if default happens in } (L,L+H],\\
		0, & \text{otherwise.}
	\end{cases}
	\tag{8}
\end{equation}
Landmarking converts the survival prediction problem into a sequence of aligned prediction tasks indexed by $L$. This provides (i) a fair comparison across models at identical information sets, (ii) robustness under drift by conditioning on “current” portfolio information, and (iii) a direct route to multi-horizon evaluation.

\begin{figure}[!htbp]
	\centering
	\resizebox{0.96\textwidth}{!}{
	\begin{tikzpicture}[x=1cm,y=1cm,>=stealth,every node/.style={font=\small}]
		\draw[->,thick] (0,0) -- (13.0,0) node[right]{Loan age};
		\draw[thick] (6.4,-0.25) -- (6.4,3.0);
		\draw[thick] (10.7,-0.25) -- (10.7,1.7);
		\node[above] at (6.4,3.0) {Landmark $L$};
		\node[above] at (10.7,1.7) {$L+H$};

		\draw[fill=gray!15,draw=black] (0.5,0.45) rectangle (6.4,1.35);
		\node[align=center] at (3.45,0.90) {Observed covariates\\ $X_i(L)$ from history up to $L$};

		\draw[fill=blue!10,draw=black] (0.5,1.65) rectangle (6.4,2.55);
		\node[align=center] at (3.45,2.10) {Lookback for behavioural marker\\ $\hat{m}_i(L)$ from $BD\_pct_i(t)$, $t\leq L$};

		\draw[fill=orange!15,draw=black] (6.4,0.45) rectangle (10.7,1.35);
		\node[align=center] at (8.55,0.90) {Prediction horizon\\ $(L,L+H]$};

		\draw[fill=orange!10,draw=black,rounded corners] (6.8,-1.25) rectangle (10.3,-0.35);
		\node[align=center] at (8.55,-0.80) {Target equals 1\\ if default occurs};

		\draw[fill=green!10,draw=black,rounded corners] (2.0,-2.55) rectangle (8.8,-1.65);
		\node[align=center] at (5.4,-2.10) {Sample includes loans still at risk at $L$\\ no information after $L$ enters prediction};

		\draw[->,thick] (8.55,0.45) -- (8.55,-0.35);
		\draw[->,thick] (6.4,-0.25) -- (6.4,-1.65);
	\end{tikzpicture}}
	\caption{Landmark sampling scheme, lookback information for the longitudinal marker, observation point and prediction horizon.}
	\label{fig:landmark_sampling_scheme}
\end{figure}

\subsection{Discrete-Time Hazard (Logistic survival model)}
Given the landmark target (8), we model the conditional probability of default within the next horizon as a discrete-time hazard quantity. Define:
\begin{equation}
	h_L \;=\;\Pr\!\left(Y^{(L,H)}=1\mid X(L)\right).
	\tag{9}
\end{equation}
Then we adopt a logistic link:
\begin{equation}
	\text{logit}(h_L)=\alpha_0+\alpha^\top X(L)+\gamma\cdot \hat{m}_i(L),
	\tag{10}
\end{equation}
where $\alpha_0$ is an intercept, $\alpha$ is a coefficient vector and $\gamma$ links the longitudinal behavioural state to default risk. To further explain, $\gamma$ indicates how sensitive monthly default probability is to how much the borrower’s actual behaviour has diverged from expected payment behaviour. It represents the strength of the endogenous feedback loop between borrower behaviour and default timing.

The logistic function yields:
\begin{equation}
	h_L=\frac{\exp\!\left(\alpha_0+\alpha^\top X(L)+\gamma\hat{m}_i(L)\right)}
	{1+\exp\!\left(\alpha_0+\alpha^\top X(L)+\gamma\hat{m}_i(L)\right)}.
	\tag{11}
\end{equation}
Model (10)--(11) is conceptually equivalent to survival modelling in discrete time, but it is fitted as a standard probabilistic classifier on landmark-constructed samples, which is empirically stable under data drift and class imbalance when combined with appropriate weighting and regularisation \citep{botha2025approaches,breeden2022multihorizon}.

\subsection{Landmark One-Hot Encoding (LM): adjusting for temporal non-stationarity}
Different landmark months correspond to systematically different risk distributions due to seasoning effects, survivorship and macro regime changes. To adjust the average risk level at each month $L$, we add landmark indicators via one-hot encoding:
\begin{equation}
	Z_L=\text{OneHot}(L).
	\tag{12}
\end{equation}
The hazard model becomes:
\begin{equation}
	\text{logit}(h_L)=\alpha_0+\alpha^\top X(L)+\gamma\cdot \hat{m}_i(L)+\delta^\top Z_L,
	\tag{13}
\end{equation}
where $\delta$ provides landmark-specific intercept adjustments. Equation (13) can also be read as a landmark-specific baseline:
\begin{equation}
	\alpha_0(L)=\alpha_0+\delta^\top Z_L,
	\qquad\Rightarrow\qquad
	\text{logit}(h_L)=\alpha_0(L)+\alpha^\top X(L)+\gamma\hat{m}_i(L).
	\tag{14}
\end{equation}
Thus, LM explicitly captures temporal non-stationarity in the baseline hazard component without forcing the covariate effects $\alpha$ to change across landmarks. This separation reduces bias under drift, where both base rates and feature distributions may shift over time \citep{hinder2024one}.
For operational prediction on new observations, the landmark one-hot vector is defined using the same landmark grid specified during model development. If a new observation is made at a landmark month already included in the grid, the corresponding element of $Z_L$ is activated and the estimated coefficient $\delta_L$ is used directly. If prediction is required at an intermediate month, the observation can be assigned to the nearest pre-specified landmark or to the corresponding landmark interval. If prediction is required beyond the maximum landmark used in training, the final landmark category can be retained as a conservative temporary extrapolation. However, in practice the model should be periodically refitted when the portfolio moves into a materially new age range or macroeconomic regime.

\subsection{Isotonic Calibration (ISO): making predicted probabilities well-calibrated}
Even when discrimination is high, probability outputs may be miscalibrated because of class imbalance and data drift. To obtain well-calibrated default probabilities, we apply isotonic regression as a monotone post-hoc calibrator.

Let $p_{raw}$ denote the raw predicted probability from the LM hazard model (13):
\begin{equation}
	p_{raw}=h_L.
	\tag{15}
\end{equation}
Isotonic calibration fits a monotone mapping $f_{iso}$ so that calibrated probabilities are:
\begin{equation}
	p_{cal}=f_{iso}(p_{raw}).
	\tag{16}
\end{equation}
The mapping $f_{iso}$ is achieved by solving the monotone least-squares problem:
\begin{equation}
	f_{iso}=\arg\min_{f\in \mathcal{F}_{monotone}}
	\sum_{i=1}^{n}\left(y_i-f(p_i)\right)^2,
	\tag{17}
\end{equation}
where $\mathcal{F}_{monotone}$ is the set of non-decreasing functions on $[0,1]$. Because $f_{iso}$ is constrained to be monotone, isotonic calibration preserves the rank ordering implied by $p_{raw}$ (hence typically preserving AUC), while correcting probability distortions that harm calibration metrics such as the Brier score \citep{ojeda2023calibrating}.

\subsection{Joint LMISO model summary}
Combining (i) a balance-based longitudinal trajectory $\hat{m}_i(L)$ from (7), (ii) landmark-conditioned discrete-time hazard modelling (9)--(11), (iii) temporal baseline adjustment via LM (12)--(14), and (iv) monotone probability calibration via ISO (16)--(17), yields the joint LMISO model used throughout the experiments. The final predicted probability of default within horizon $H$ from landmark $L$ is:
\begin{equation}
	\boxed{
		\Pr\!\left(Y^{(L,H)}=1\mid X(L)\right)
		\;=\;p_{cal}
		\;=\;f_{iso}\!\left(\sigma\!\left(\alpha_0+\alpha^\top X(L)+\gamma\hat{m}_i(L)+\delta^\top Z_L\right)\right)
	}
	\tag{18}
\end{equation}
where $\sigma(\cdot)$ denotes the logistic function.

\section{Data and Experimental Settings}

\subsection{Data Description and Preprocessing}
In this study we adopt the public Single Family Loan-Level Dataset updated by June 30, 2025 from Freddie Mac. For the purpose of reducing computational cost, we use the randomly selected sample dataset of 50,000 loans for each year instead of all loans in one year. There are two kinds of datasets: origination data (including time-fixed covariates) and monthly performance data (including time-varying variables). Both are simultaneously used for prediction and experiments are conducted on drifted datasets from year 2000, 2010, 2020 and 2021.

For origination data, the following static predictors are kept as displayed in Table~\ref{tab:description_orig}.
\begin{table}[!htbp]
	\centering
	\caption{Description of static covariates in origination data}
	\label{tab:description_orig}
	\begin{tabular}{p{3.5cm} p{10cm}}
		\hline
		\textbf{Covariate} & \textbf{Description} \\
		\hline
		\textit{CreditScore} &
		A number, prepared by third parties, summarizing the borrower’s creditworthiness. Generally, the credit score disclosed is the score known at the time of acquisition and is the score used to originate the mortgage. \\
		
		\textit{Occupancy} &
		Denotes whether the mortgage type is owner occupied, second home, or investment property. \\
		
		\textit{DTI} &
		The debt to income ratio. \\
		
		\textit{OrigUPB} &
		The original unpaid principal balance of the mortgage on the note date. \\
		
		\textit{OrigLTV} &
		The original loan to value. \\
		
		\textit{OrigInterestRate} &
		The interest rate of the loan as stated on the note at the time the loan was originated. \\
		
		\textit{LoanPurpose} &
		Indicates whether the mortgage loan is a Cash-out Refinance mortgage, No Cash-out Refinance mortgage, or a Purchase mortgage. \\
		
		\textit{OrigLoanTerm} &
		A calculation of the number of scheduled monthly payments of the mortgage based on the First Payment Date and Maturity Date. \\
		
		\textit{NumBorrowers} &
		The number of Borrower(s) who are obligated to repay the mortgage note secured by the mortgaged property. \\
		\hline
	\end{tabular}
\end{table}
For monthly performance data, in addition to those indicating the date, the following time-varying variables are utilized as illustrated in Table~\ref{tab:description_serv}.
\begin{table}[!htbp]
	\centering
	\caption{Description of time-varying variables in monthly performance data}
	\label{tab:description_serv}
	\begin{tabular}{p{3.5cm} p{10cm}}
		\hline
		\textbf{Variable} & \textbf{Description} \\
		\hline
		\textit{CurAct\_UPB} &
		The Current Actual UPB reflects the mortgage ending balance as reported by the servicer for the corresponding monthly reporting period. \\
		
		\textit{CurLoanDel} &
		The current loan delinquency status is a value corresponding to the number of months the borrower is delinquent. \\
		
		\textit{LoanAge} &
		The number of scheduled payments from the time the loan was originated up to and including the current period. \\
		
		\textit{ZeroBalCode} &
		The zero balance code is a code indicating the reason the loan's balance was reduced to zero. \\
		
		\textit{CurIntRate} &
		Reflects the current interest rate on the mortgage note, taking into account any loan modifications. \\
		
		\textit{CNIB\_UPB} &
		The current non-interest bearing portion of the unpaid principal balance for a given mortgage. \\
		
		\textit{ELTV} &
		The estimated loan to value is a ratio indicating current LTV based on the estimated current value of the property obtained through Freddie Mac’s Automated Valuation Model (AVM). \\
		
		\textit{ASSISTANCE\_CODE} &
		Regardless of delinquency status, the type of assistance plan that the borrower is enrolled in that provides temporary mortgage payment relief or an opportunity to cure a mortgage delinquency over a defined period. F = Forbearance; R = Repayment; T = Trial Period; Null = No workout plan or 
		not applicable. \\
		\hline
	\end{tabular}
\end{table}
Loans that cannot be linked by $LoanSeqNum$ (LOAN SEQUENCE NUMBER) in two datasets are removed. Invalid values are coded as missing when they violate variable definitions or mortgage-domain constraints. Specifically, negative balances, negative loan terms, non-positive original unpaid balances, implausible loan terms above 1000 months, and interest-rate or loan-to-value observations outside economically meaningful ranges are treated as missing. We avoid winsorising these observations because extreme values may reflect data quality problems rather than genuine borrower behaviour. Prepayment is not modelled as a separate competing risk in the present analysis, which keeps the focus on default prediction. $CurLoanDel$ is identified as the target variable, and a default is considered to have occurred when its value is not 0. The threshold $CurLoanDel\neq0$ is intentionally adopted as an early-delinquency definition rather than as a conventional default definition based on the Basel criterion. The model is evaluated on whether it detects deterioration early enough for intervention, matching the early-warning objective of portfolio monitoring. This definition also yields a less sparse event process and permits meaningful F1-score comparison under grouped cross-validation. To assess the implication of a common default threshold, Appendix Tables~\ref{tab:ge3_sudden}--\ref{tab:ge3_recurring} report a sensitivity analysis using $CurLoanDel\geq3$ for the 2020 drift scenarios. The results show that the 90-days-plus definition significantly reduces positive cases and depresses F1 for the hazard-based models, even when AUC remains moderate. This confirms that the two thresholds represent different operational tasks.

\subsection{Data Drift Simulation}
In this study, concept drifts (sudden, incremental and recurring) and label drift are conducted simultaneously and independent of each other. Only $CurAct\_UPB$, $CurIntRate$, $ELTV$ and $CurLoanDel$ are changed to simulate shifts in the mapping from time-varying predictors to default. We implement drift via time-indexed parametric schedules (Figure~\ref{fig:drift_schedules}) over the known observation month $t=1,\dots,T$: sudden drift is introduced as a one-time structural break at $t_s=\lfloor T/3\rfloor$, where $CurIntRate$ is shifted by $+1$, $ELTV$ is scaled by $1.2$ for $t\ge t_s$, and the first post-break $CurAct\_UPB$ observation per loan is reduced by $5\%$; incremental drift is imposed by linearly ramping the perturbation strength from $t_s=\lfloor T/3\rfloor$ to $t_e=\lfloor 2T/3\rfloor$, yielding $CurIntRate \leftarrow CurIntRate + 1.5\,\tau(t)$, $ELTV \leftarrow ELTV\cdot(1+0.15\,\tau(t))$, and $CurAct\_UPB \leftarrow CurAct\_UPB\cdot(1-0.09\,\tau(t))$, where $\tau(t)\in[0,1]$ increases linearly on $[t_s,t_e]$ and is clipped outside; recurring drift is generated as a 12-month sinusoidal cycle from the start of data collection, with $CurIntRate \leftarrow CurIntRate + 0.5\sin(2\pi t/12)$, $ELTV \leftarrow ELTV\cdot(1+0.05\sin(2\pi t/12+\pi/6))$, and $CurAct\_UPB \leftarrow CurAct\_UPB\cdot\bigl(1-0.02(0.5+0.5\sin(2\pi t/12+\pi/3))\bigr)$. Label drift is implemented by manipulating the monthly prevalence of delinquency $p(t)=\mathbb{P}(CurLoanDel\neq 0)$ through probabilistic label flipping within each month: for sudden drift $p(t)$ shifts from $0.025$ to $0.10$ at $t_s$, for incremental drift $p(t)$ increases linearly from $0.025$ to $0.12$ on $[t_s,t_e]$, and for recurring drift it oscillates as $0.06 + 0.035\sin(\cdot)$. 

These controlled, interpretable drift regimes are designed to reflect economically meaningful forms of mortgage portfolio instability. Sudden drift approximates abrupt structural breaks, such as interest-rate shocks, policy interventions, or sharp house-price corrections. Incremental drift represents gradual deterioration or improvement in borrower affordability and collateral values during prolonged macroeconomic transitions. Recurring drift mimics seasonal or cyclical patterns in repayment behaviour and housing-market conditions. The perturbed variables are selected because they are directly linked to repayment burden, collateral risk, and delinquency formation: $CurAct\_UPB$ reflects the realised balance path, $CurIntRate$ captures financing cost, $ELTV$ proxies collateral deterioration, and $CurLoanDel$ defines the observed delinquency state. Different simulation schemes are adopted accordingly because the data from different years in Freddie Mac vary considerably. The drift transformations are applied to the full time-indexed monthly performance panel before the grouped cross-validation folds are constructed by $LoanSeqNum$. Thus, drift is not restricted to the in-sample training portion only. Each loan in each training or test fold owns a full time drifted trajectory. This design evaluates whether a model trained on a representative portion of simulated data generalises to unseen loans exposed to the same type of non-stationarity.
\begin{figure}[!htbp]
	\centering
	\begin{tikzpicture}
		\begin{groupplot}[
			group style={group size=2 by 2, horizontal sep=1.5cm, vertical sep=2.0cm},
			width=0.48\textwidth,
			height=0.33\textwidth,
			xmin=1, xmax=60,
			xtick={1,12,24,36,48,60},
			xlabel={Month index $t$},
			grid=major,
			legend style={at={(0.5,-0.25)},anchor=north,legend columns=3},
			legend cell align=left,
			]
			
			\nextgroupplot[
			title={$\,\Delta CurIntRate(t)$ (additive)},
			ylabel={Additive shift},
			]
			\addplot[thick, blue, mark=none] coordinates {(1,0) (20,0) (20,1) (60,1)};
			\addplot[thick, orange, dashed, mark=none] coordinates {(1,0) (20,0) (40,1.5) (60,1.5)};
			\addplot[thick, green!60!black, dashdotted, mark=none, domain=1:60, samples=240]
			{0.5*sin(deg(2*pi*x/12))};
			
			\nextgroupplot[
			title={$\,m_{ELTV}(t)$ (multiplicative)},
			ylabel={Scale factor},
			ymin=0.8, ymax=1.3
			]
			\addplot[thick, blue, mark=none] coordinates {(1,1) (20,1) (20,1.2) (60,1.2)};
			\addplot[thick, orange, dashed, mark=none] coordinates {(1,1) (20,1) (40,1.15) (60,1.15)};
			\addplot[thick, green!60!black, dashdotted, mark=none, domain=1:60, samples=240]
			{1 + 0.05*sin(deg(2*pi*x/12 + pi/6))};
			
			\nextgroupplot[
			title={$\,m_{UPB}(t)$ (multiplicative)},
			ylabel={Scale factor},
			ymin=0.9, ymax=1.02
			]
			\addplot[thick, blue, mark=*] coordinates {(1,1) (19,1) (20,0.95) (21,1) (60,1)};
			\addplot[thick, orange, dashed, mark=none] coordinates {(1,1) (20,1) (40,0.91) (60,0.91)};
			\addplot[thick, green!60!black, dashdotted, mark=none, domain=1:60, samples=240]
			{1 - 0.02*(0.5 + 0.5*sin(deg(2*pi*x/12 + pi/3)))};
			
			\nextgroupplot[
			title={$\,p(t)=\mathbb{P}(CurLoanDel\neq 0)$},
			ylabel={Prevalence},
			ymin=0.0, ymax=0.15,
			scaled y ticks=false,
			ytick={0,0.05,0.10,0.15},
			yticklabel style={/pgf/number format/fixed, /pgf/number format/precision=2},
			]
			\addplot[thick, blue, mark=none] coordinates {(1,0.025) (20,0.025) (20,0.10) (60,0.10)};
			\addplot[thick, orange, dashed, mark=none] coordinates {(1,0.025) (20,0.025) (40,0.12) (60,0.12)};
			\addplot[thick, green!60!black, dashdotted, mark=none, domain=1:60, samples=240]
			{0.06 + 0.035*sin(deg(2*pi*x/12 - pi/4))};
			
			\legend{Sudden,Incremental,Recurring}
		\end{groupplot}
	\end{tikzpicture}
	\caption{Parametric drift schedules used for simulation (schematic). The month index spans the first 60 months for illustration; $t_s=20$ and $t_e=40$ correspond to one-third and two-thirds of the span, respectively.}
	\label{fig:drift_schedules}
\end{figure}

We next assess the drift level of simulated datasets. For numeric variables: we compute per-loan median month-over-month absolute change normalized by the variable’s global Interquartile Range (IQR). Thresholds of drift level are [None=(0, 0.1); Slight=(0.1, 0.3); Moderate=(0.3, 0.7); Severe>0.7]. For categorical variables: we compute per-loan state-change rate over time. Thresholds of drift level are [None=(0, 0); Slight=(0, 0.1); Moderate=(0.1, 0.3); Severe=(0.3, 1.0)]. We take the data from year 2020 as an example to show the effect of data drift. Figure~\ref{fig:drift_level_comparison} reports, for each drift setting, the distribution of loans across drift-severity levels for $CurAct\_UPB$, $CurIntRate$, $ELTV$, and $CurLoanDel$ using stacked bars, so that both the overall drift prevalence and the severity mix can be compared at a glance. In particular, a rightward shift of bar mass toward ``Moderate'' and ``Severe'' indicates a stronger and more widespread deviation of the variable’s temporal dynamics under the corresponding simulated drift regime.
\begin{figure}[!htbp]
	\centering
	\begin{tikzpicture}
		\begin{groupplot}[
			group style={group size=2 by 2, horizontal sep=1.4cm, vertical sep=2.0cm},
			ybar stacked,
			width=0.48\textwidth,
			height=0.34\textwidth,
			ymin=0, ymax=100,
			ylabel={Percentage (\%)},
			symbolic x coords={Original,Sudden,Incremental,Recurring},
			xtick=data,
			xticklabels={Original,Sudden,Incremental,Recurring},
			x tick label style={rotate=25, anchor=east},
			grid=major,
			legend style={at={(0.5,-0.25)},anchor=north,legend columns=5},
			]
			
			\nextgroupplot[title={$CurAct\_UPB$}]
			\addplot coordinates {(Original,60.89) (Sudden,37.61) (Incremental,45.61) (Recurring,17.87)};
			\addplot coordinates {(Original,30.14) (Sudden,21.17) (Incremental,37.44) (Recurring,12.08)};
			\addplot coordinates {(Original,6.31)  (Sudden,5.70)  (Incremental,13.06) (Recurring,8.31)};
			\addplot coordinates {(Original,2.62)  (Sudden,35.48) (Incremental,3.84)  (Recurring,61.69)};
			\addplot coordinates {(Original,0.04)  (Sudden,0.04)  (Incremental,0.05)  (Recurring,0.05)};
			
			\nextgroupplot[title={$CurIntRate$}]
			\addplot coordinates {(Original,0.00) (Sudden,0.00) (Incremental,1.88) (Recurring,98.39)};
			\addplot coordinates {(Original,0.00) (Sudden,0.00) (Incremental,0.54) (Recurring,1.57)};
			\addplot coordinates {(Original,0.00) (Sudden,0.00) (Incremental,0.01) (Recurring,0.00)};
			\addplot coordinates {(Original,99.96) (Sudden,99.96) (Incremental,97.53) (Recurring,0.00)};
			\addplot coordinates {(Original,0.04) (Sudden,0.04) (Incremental,0.04) (Recurring,0.04)};
			
			\nextgroupplot[title={$ELTV$}]
			\addplot coordinates {(Original,24.77) (Sudden,22.81) (Incremental,32.80) (Recurring,57.77)};
			\addplot coordinates {(Original,69.18) (Sudden,70.97) (Incremental,60.92) (Recurring,37.38)};
			\addplot coordinates {(Original,0.59) (Sudden,1.12)  (Incremental,1.19)  (Recurring,1.61)};
			\addplot coordinates {(Original,5.43) (Sudden,1.98) (Incremental,1.97) (Recurring,0.12)};
			\addplot coordinates {(Original,0.03) (Sudden,3.12) (Incremental,3.12) (Recurring,3.12)};
			
			\nextgroupplot[title={$CurLoanDel$}]
			\addplot coordinates {(Original,1.52) (Sudden,64.20) (Incremental,61.93) (Recurring,72.08)};
			\addplot coordinates {(Original,2.78) (Sudden,1.93)  (Incremental,1.96)  (Recurring,2.45)};
			\addplot coordinates {(Original,5.60) (Sudden,1.12)  (Incremental,1.13)  (Recurring,1.12)};
			\addplot coordinates {(Original,90.06) (Sudden,0.64) (Incremental,0.61) (Recurring,0.62)};
			\addplot coordinates {(Original,0.04) (Sudden,32.11) (Incremental,34.37) (Recurring,23.73)};
			
			\legend{Severe,Moderate,Slight,None,Insufficient}
		\end{groupplot}
	\end{tikzpicture}
	\caption{Drift level comparison 2020.}
	\label{fig:drift_level_comparison}
\end{figure}

\subsection{Experimental Settings}
Primary experiments are divided into two parts: the comparison between LMISO and other popular related models, and an ablation analysis to quantify the marginal value of each component in the proposed LMISO framework. In the first part, we select four comparison objects: Cox as a survival analysis benchmark \citep{therneau2000cox}; XGBoost as a strong machine learning benchmark \citep{chen2016xgboost}; Hoeffding Adaptive Tree (HAT) and Adaptive Random Forest (ARF) as two drift-oriented benchmarks \citep{bifet2009adaptive,gomes2017adaptive}. In the second part, the baseline discrete-time hazard model is compared with further different variants. As shown in Table~\ref{tab:variants_M1}, most mechanisms are elaborated in Methodology except M1-TD and M1-IW.

M1-TD uses the same landmark target and covariate set as M1-Joint, but modifies the contribution of each training observation through a time-decay weight. Let $L_j$ denote the landmark associated with training observation $j$, and let $L_{\max}$ be the largest landmark observed in the corresponding training fold. The weight assigned to observation $j$ is
\begin{equation}
	w^{TD}_j=\exp\left[-\gamma\frac{L_{\max}-L_j}{H}\right],
	\tag{19}
\end{equation}
where $H$ is the prediction horizon and $\gamma=0.20$ in the experiments. More recent landmark observations therefore receive larger weights, while earlier landmark observations are downweighted. The weighted logistic objective is
\begin{equation}
	\min_{\theta}
	-\sum_{j=1}^{n} w^{TD}_j
	\left[y_j\log p_j+(1-y_j)\log(1-p_j)\right],
	\tag{20}
\end{equation}
where $p_j=\Pr(Y_j=1\mid X_j)$ is the predicted horizon-default probability. This design is intended to improve robustness under incremental drift, where recent observations may be more representative of the current risk environment.

M1-IW also retains the same landmark target and features as M1-Joint, but reweights training observations according to their similarity to the test-fold covariate distribution. For each cross-validation fold, a domain classifier is trained to distinguish training observations from test observations. Let $D_j=1$ indicate that observation $j$ belongs to the test fold and $D_j=0$ indicate that it belongs to the training fold. The domain classifier estimates
\begin{equation}
	s_j=\Pr(D_j=1\mid X_j).
	\tag{21}
\end{equation}
The corresponding importance weight for training observation $j$ is
\begin{equation}
	w^{IW}_j=\frac{s_j}{1-s_j}.
	\tag{22}
\end{equation}
To avoid instability from extreme density-ratio estimates, weights are clipped at $10\times\operatorname{median}(w^{IW})$. The final hazard model is then fitted by minimising the weighted logistic loss
\begin{equation}
	\min_{\theta}
	-\sum_{j=1}^{n} w^{IW}_j
	\left[y_j\log p_j+(1-y_j)\log(1-p_j)\right].
	\tag{23}
\end{equation}
This approach follows the standard covariate-shift correction principle that training observations more similar to the evaluation distribution should contribute more strongly to parameter estimation.
\begin{table}[!htbp]
	\centering
	\caption{Variants of Model One (M1)}
	\label{tab:variants_M1}
	\begin{tabular}{p{3.5cm} p{10cm}}
		\hline
		\textbf{Variant} & \textbf{Description} \\
		\hline
		\textit{M1} &
		The baseline discrete-time hazard model without the longitudinal marker. \\
		
		\textit{M1-Joint} &
		Extends M1 by integrating the longitudinal behavioural design derived from balance mechanics. \\
		
		\textit{M1-LM} &
		Augments M1-Joint with landmark one-hot encoding to model temporal heterogeneity across landmark ages. \\
		
		\textit{M1-TD} &
		Applies exponential time-decay (TD) by the distance from a training row's landmark to the maximum landmark in the training fold so that more recent landmarking observations contribute more strongly to parameter estimation. \\
		
		\textit{M1-IW} &
		Uses covariate-shift importance weighting (IW) obtained from a train–test domain classifier to correct for covariate shift by upweighting training samples that look most like the held-out fold's covariate distribution. \\
		
		\textit{M1-LMISO} &
		Combines M1-LM with isotonic calibration, which applies a monotone mapping to improve probability calibration under drift. \\
		\hline
	\end{tabular}
\end{table}

The number of landmarks is determined by the duration of specific datasets. Typically, predictions are made quarterly starting from the twelfth month and the horizon $H$ is set to be 12. For data from the same year, all models are trained and evaluated using the same landmark-based samples and prediction targets. M1-LMISO is implemented in Python using scikit-learn. The discrete-time hazard component is fitted with Logistic Regression setting \textit{max\_iter = 3000}, and class-weight balancing to address the rare-event structure of default. The model uses the pre-specified covariate set available at each landmark time: static origination variables (\textit{CreditScore}, \textit{DTI}, \textit{OrigLTV}, \textit{OrigInterestRate}, \textit{OrigLoanTerm}, and \textit{NumBorrowers}), normalised loan age, the fitted longitudinal balance-deviation marker, one-hot encoded categorical variables (\textit{Occupancy} and \textit{LoanPurpose}), and landmark one-hot indicators. No automatic variable selection is applied; variables are selected a priori based on mortgage-domain relevance, availability before the prediction horizon, and consistency with the proposed longitudinal-hazard design. After fitting the logistic hazard model, raw training-fold probabilities are passed to Isotonic Regression with \textit{out\_of\_bounds = ``clip''}; the fitted monotone calibration mapping is then applied to the corresponding test-fold probabilities. Hyperparameters and other detailed settings are accessible on GitHub through this link [https://doi.org/10.5281/zenodo.18392896].

Model evaluation is carried out using five-fold grouped cross-validation (CV), where grouping is by \textit{LoanSeqNum}. This ensures that all observations from the same loan appear either in the training fold or in the test fold, but never in both, thereby preventing information leakage across repeated monthly observations from the same borrower. The drift transformation is applied to the full monthly panel before CV. Therefore, the train and test folds differ by loan identities rather than by an intentionally imposed train-versus-test temporal regime split. Both training and test loans are exposed to the same simulated drift mechanism and contain full drifted trajectories.

This design is appropriate for the purpose of the empirical analysis. The aim is not to reproduce a single deployment setting in which a model is trained on one historical regime and then evaluated on a future regime. Instead, the aim is to compare model behaviour under controlled non-stationary data-generating processes. In this setting, drift is represented by time-indexed changes in covariates and labels within the monthly loan panel, for example through sudden, incremental, and recurring changes in repayment burden, collateral risk, and delinquency prevalence. Thus, non-stationarity is present within the data-generating process itself, while grouped CV provides out-of-sample evaluation with respect to unseen loans.

It is not necessary to make the train and test folds deliberately different in temporal distribution for this specific purpose. A train-on-early-periods and test-on-later-periods design answers a different question: how well a model extrapolates from one calendar regime to a later regime. Such an out-of-time design is valuable for external validation, but it also introduces additional confounding from cohort composition, vintage effects, macroeconomic conditions, censoring patterns, and the distribution of available landmark ages. These factors make it harder to isolate the effect of the simulated drift shape itself. By contrast, grouped CV keeps the simulated drift environment, landmark construction, prediction horizon, and target definition comparable across models, leading performance differences to mainly reflect the modelling mechanisms rather than a particular temporal split.

Accordingly, the present protocol should be considered as a grouped out-of-loan validation under controlled non-stationary scenarios. It evaluates whether each model generalises to unseen loans that evolve under the same designed drift process.

The same folds are used for all models to ensure comparability. Evaluation metrics are computed at the fold level and then aggregated by reporting the mean and standard deviation (SD) across folds. Given that Freddie Mac loan datasets are class-imbalanced, we choose the area under the receiver operating characteristic curve (AUC) to assess the discrimination of model performance. Calibration is measured by the Brier score, which captures the mean squared error between predicted probabilities and observed outcomes. F1 score represents models' quality of detecting positive-class samples.

\section{Results and Analysis}
This section reports model performance under three simulated drift environments: sudden drift, incremental drift and recurring drift. The results are organised into two parts as mentioned in Experimental Settings.

\subsection{Comparison with benchmark models}
We first compare M1-LMISO against four previously mentioned benchmarks.
\begin{table}[!htbp]
	\centering
	\caption{Benchmark comparison under \textbf{sudden drift} (mean (SD), 5-fold grouped CV).}
	\label{tab:benchmark_sudden}
	\begin{tabular}{lccc}
		\toprule
		Model & AUC & Brier & F1 \\
		\midrule
		M1-LMISO & 0.812 (0.004) & 0.102 (0.002) & 0.924 (0.001) \\
		Cox      & 0.571 (0.004) & 0.372 (0.003) & 0.346 (0.003) \\
		XGBoost  & 0.794 (0.005) & 0.126 (0.003) & 0.890 (0.003) \\
		HAT      & 0.622 (0.041) & 0.135 (0.004) & 0.921 (0.002) \\
		ARF      & 0.533 (0.006) & 0.140 (0.003) & 0.921 (0.002) \\
		\bottomrule
	\end{tabular}
\end{table}

\begin{table}[!htbp]
	\centering
	\caption{Benchmark comparison under \textbf{incremental drift} (mean (SD), 5-fold grouped CV).}
	\label{tab:benchmark_incremental}
	\begin{tabular}{lccc}
		\toprule
		Model & AUC & Brier & F1 \\
		\midrule
		M1-LMISO & 0.836 (0.003) & 0.131 (0.001) & 0.876 (0.001) \\
		Cox      & 0.666 (0.002) & 0.340 (0.001) & 0.326 (0.002) \\
		XGBoost  & 0.826 (0.002) & 0.151 (0.001) & 0.847 (0.001) \\
		HAT      & 0.431 (0.040) & 0.229 (0.006) & 0.865 (0.002) \\
		ARF      & 0.510 (0.009) & 0.222 (0.002) & 0.865 (0.002) \\
		\bottomrule
	\end{tabular}
\end{table}

\begin{table}[!htbp]
	\centering
	\caption{Benchmark comparison under \textbf{recurring drift} (mean (SD), 5-fold grouped CV).}
	\label{tab:benchmark_recurring}
	\begin{tabular}{lccc}
		\toprule
		Model & AUC & Brier & F1 \\
		\midrule
		M1-LMISO & 0.696 (0.006) & 0.115 (0.005) & 0.923 (0.004) \\
		Cox      & 0.518 (0.005) & 0.222 (0.002) & 0.003 (0.003) \\
		XGBoost  & 0.662 (0.008) & 0.144 (0.004) & 0.882 (0.004) \\
		HAT      & 0.540 (0.007) & 0.128 (0.005) & 0.923 (0.003) \\
		ARF      & 0.518 (0.008) & 0.129 (0.005) & 0.923 (0.003) \\
		\bottomrule
	\end{tabular}
\end{table}

Tables~\ref{tab:benchmark_sudden}--\ref{tab:benchmark_recurring} compare M1-LMISO with four common benchmarks under sudden, incremental and recurring drift. Across all drift regimes, LMISO delivers the most favourable combination of discrimination (AUC), calibration (Brier) and thresholded performance (F1). The Cox model exhibits the weakest overall performance with continuously poor AUC, high Brier scores and extremely low F1 scores. This suggests that its proportional hazards structure fails to capture the combined effects of temporal heterogeneity and dynamic behavioural signals present under data drift. Furthermore, changes in borrower composition over loan age and time-varying conditions can alter both the baseline risk level and covariate effects, violating proportional hazard assumptions and producing considerable probability distortion when evaluated on held-out folds.

XGBoost is the only comparison model that is competitive across all metrics and drifts. Main hyperparameters are fixed across all drift scenarios to avoid scenario-specific overfitting: maximum tree depth is set to 15, the learning rate to 0.05, subsample and column-sample ratios to 0.8, and the maximum number of boosting rounds to 500 with early stopping after 25 rounds. Class imbalance is handled by setting \textit{scale\_pos\_weight} equal to the ratio of negative to positive cases in the training fold. The fixed configuration is selected before comparing drift scenarios so that XGBoost serves as a common benchmark rather than a scenario-tuned competitor. The relatively large tree depth allows nonlinear interactions in the landmark features, the low learning rate and early stopping limit overfitting, and the subsample and column-sample ratios reduce variance. The only fold-specific quantity is \textit{scale\_pos\_weight}, which is recomputed from the training fold because default prevalence changes across folds and drift settings. This pattern is consistent with the fact that boosting is highly flexible for fitting complex decision boundaries. However, the lack of an inherent temporal structure limits its effectiveness when processing non-stationary data streams.

The two drift-oriented online learners, HAT and ARF, show an even more pronounced mismatch: in several settings they achieve relatively high F1 but only weak AUC, indicating that they may be regarded as coarse threshold classifiers that perform reasonably at a particular operating point while failing to rank risk reliably across loans. This is expected because these methods are primarily designed for online point classification in streaming environments rather than for survival-style forecasting with landmark conditioning and horizon-based targets. Moreover, the learning signal for default is highly imbalanced and delayed in credit risk. Drift detectors and adaptive replacement mechanisms therefore overreact to noise or underreact to slow shifts, which can yield unstable probability estimates and poor ranking. LMISO avoids these pitfalls by adopting (i) a survival-aligned landmark target; (ii) a behaviour-driven longitudinal summary that remains informative under covariate drift; (iii) explicit probability calibration that stabilises outputs under changing base rates.

In summary, achieving robust predictive performance under concept drift requires various mechanisms that go beyond standard modelling choices. The superiority of LMISO indicates that robustness in mortgage default prediction depends on jointly addressing behavioural dynamics, temporal heterogeneity and probability calibration under shift, which these benchmarks cannot capture simultaneously.

\subsection{Ablation analysis of LMISO}
The ablation analysis evaluates six models as shown in the tables below.
\begin{table}[!htbp]
	\centering
	\caption{Ablation results under \textbf{sudden drift} (mean (SD), 5-fold grouped CV).}
	\label{tab:ablation_sudden}
	\begin{tabular}{lccc}
		\toprule
		Model & AUC & Brier & F1 \\
		\midrule
		M1       & 0.699 (0.003) & 0.220 (0.001) & 0.773 (0.004) \\
		M1-Joint & 0.805 (0.004) & 0.186 (0.000) & 0.798 (0.001) \\
		M1-LM    & 0.812 (0.004) & 0.182 (0.000) & 0.807 (0.001) \\
		M1-TD    & 0.806 (0.004) & 0.105 (0.002) & 0.921 (0.002) \\
		M1-IW    & 0.806 (0.004) & 0.105 (0.002) & 0.921 (0.002) \\
		M1-LMISO & 0.812 (0.004) & 0.102 (0.002) & 0.924 (0.001) \\
		\bottomrule
	\end{tabular}
\end{table}

\begin{table}[!htbp]
	\centering
	\caption{Ablation results under \textbf{incremental drift} (mean (SD), 5-fold grouped CV).}
	\label{tab:ablation_incremental}
	\begin{tabular}{lccc}
		\toprule
		Model & AUC & Brier & F1 \\
		\midrule
		M1       & 0.773 (0.004) & 0.195 (0.002) & 0.785 (0.002) \\
		M1-Joint & 0.831 (0.003) & 0.172 (0.001) & 0.810 (0.001) \\
		M1-LM    & 0.836 (0.003) & 0.168 (0.001) & 0.814 (0.001) \\
		M1-TD    & 0.831 (0.003) & 0.134 (0.001) & 0.877 (0.001) \\
		M1-IW    & 0.831 (0.003) & 0.135 (0.001) & 0.876 (0.002) \\
		M1-LMISO & 0.836 (0.003) & 0.131 (0.001) & 0.876 (0.001) \\
		\bottomrule
	\end{tabular}
\end{table}

\begin{table}[!htbp]
	\centering
	\caption{Ablation results under \textbf{recurring drift} (mean (SD), 5-fold grouped CV).}
	\label{tab:ablation_recurring}
	\begin{tabular}{lccc}
		\toprule
		Model & AUC & Brier & F1 \\
		\midrule
		M1       & 0.623 (0.003) & 0.239 (0.001) & 0.718 (0.003) \\
		M1-Joint & 0.691 (0.006) & 0.224 (0.000) & 0.736 (0.002) \\
		M1-LM    & 0.696 (0.006) & 0.222 (0.000) & 0.744 (0.002) \\
		M1-TD    & 0.691 (0.006) & 0.116 (0.005) & 0.923 (0.004) \\
		M1-IW    & 0.691 (0.006) & 0.116 (0.005) & 0.923 (0.004) \\
		M1-LMISO & 0.696 (0.006) & 0.115 (0.005) & 0.923 (0.004) \\
		\bottomrule
	\end{tabular}
\end{table}

Tables~\ref{tab:ablation_sudden}--\ref{tab:ablation_recurring} show three consistent patterns. Firstly, incorporating the longitudinal behavioural marker produces the greatest enhancement in discrimination: M1-Joint increases AUC over M1 in all drift settings (e.g., from 0.699 to 0.805 under sudden drift, and from 0.623 to 0.691 under recurring drift). Brier and F1 scores also improve modestly. This indicates that the balance-mechanics-based longitudinal design captures a risk-relevant behavioural signal that is not recoverable from the hazard model alone. Intuitively, the marker summarises persistent deviation and trend effects in repayment behaviour, which remain informative even when the marginal distribution of balances drifts.

This finding is broadly in line with dynamic survival and joint-modelling studies. In credit risk, \citet{medina2023joint} show that modelling repeated credit-process information jointly with default timing improves default prediction relative to specifications that treat borrower information more statically. More generally, \citet{rizopoulos2014combining} and \citet{andrinopoulou2021reflection} demonstrate that longitudinal markers enhance dynamic survival prediction because they update the subject-specific risk profile as new measurements arrive. In our setting, the per-loan regression contributes to this gain by transforming the noisy monthly balance deviation series into a low-dimensional behavioural state, $\hat{m}_i(L)$, evaluated at each landmark. This state captures both the borrower-specific level of deviation from the scheduled amortisation path and its deterioration or improvement over loan age, while filtering temporary month-to-month fluctuations and reducing scale differences across loans with different original balances and terms. Therefore, the enhancement from M1 to M1-Joint reflects the introduction of a behaviourally normalised trajectory, rather than the simple addition of another time-varying balance variable. A hazard model with lagged or rolling-window values of \textit{CurAct\_UPB}, \textit{CurIntRate} and \textit{ELTV} captures part of the recent repayment history, but this specification remains higher-dimensional, more collinear and less directly comparable across loans. Furthermore, raw lagged covariates do not distinguish normal scheduled amortisation from borrower-specific repayment deviation. Comparable gains therefore require past covariates to be engineered into economically meaningful trajectory summaries, whereas raw lookback variables alone provide a less stable and less portable signal under drift.

Secondly, landmark one-hot encoding moderately strengthens discrimination rather than calibration. Across all drift types, M1-LM achieves the highest (or tied-highest) AUC within the ablation family, suggesting that a landmark-specific baseline risk term effectively accounts for lifecycle heterogeneity and survivorship effects. By conditioning the intercept on landmark age, the model avoids conflating early-stage and late-stage risk regimes, which is especially important when drift changes the portfolio composition over time \citep{gama2014survey}.

Thirdly, calibration changes are driven by drift-aware reweighting and isotonic calibration, with LMISO providing the most robust balance between discrimination and calibration. Under sudden drift (Table~\ref{tab:ablation_sudden}), M1-TD and M1-IW sharply reduce Brier (from 0.186 in M1-Joint to about 0.105) and increase F1 to above 0.92, corresponding to the role of reweighting methods in correcting fold-level covariate imbalance between training and test loans. However, these approaches do not greatly raise AUC beyond M1-Joint. In contrast, M1-LMISO combines the discrimination benefit of landmark encoding (AUC $\approx$ 0.812) with the calibration benefit of isotonic mapping (Brier 0.102), yielding the strongest overall profile.

The same decomposition holds under incremental drift (Table~\ref{tab:ablation_incremental}). M1-LMISO matches the best AUC (0.836) and achieves the lowest Brier (0.131). This is consistent with the monotonic property of isotonic calibration: it preserves ranking (hence AUC) while correcting probability distortion (hence Brier). Under recurring drift (Table~\ref{tab:ablation_recurring}), where non-stationarity is persistent and cyclic, M1-LMISO maintains the strongest calibration (Brier 0.115) while also achieving the highest AUC (0.696) and F1 (0.923). Notably, recurring drift produces the lowest AUC levels for all models, indicating that cyclic regime changes are intrinsically harder for ranking-based risk prediction; nevertheless, LMISO remains the most stable and accurate.

Overall, the ablation results support the intended functional roles of each component: the joint longitudinal design provides the dominant discrimination gain; landmark encoding adjusts for temporal heterogeneity and improves ranking; isotonic calibration delivers additional robustness in probability estimation, particularly under different drifts.

The experimental results analysed here are based on data from 2020, while the results for the other three years provided in Appendix show similar patterns discussed for 2020.

\section{Conclusions and Future Work}
This paper studies mortgage time-to-default prediction in non-stationary environments and shows that data drift can degrade the effectiveness of widely used survival and machine learning models to varying degrees. To address this challenge, we propose a landmark-based dynamic framework that links a balance-mechanics-driven longitudinal behavioural marker to a discrete-time hazard model and further enhances robustness by means of landmark one-hot encoding (LM) and isotonic calibration (ISO). The empirical results under three drift scenarios (sudden, incremental and recurring) provide practical evidence that modelling behavioural evolution and explicitly correcting temporal heterogeneity and probability distortion are crucial for reliable default prediction under drift. Our LMISO model outperforms the classical survival model, the strong machine learning baseline and the drift-adaptive online learners across multiple drift settings. In addition, this approach remains explainable and computationally lightweight due to the action of using logistic regression, closed-form amortisation calculations and standard linear classification with a simple calibration step. This makes it feasible for financially sensitive departments of government and enterprises when handling large mortgage portfolios.

The benchmark comparison further highlights why an integrated survival-and-drift perspective is necessary. Cox regression performs terribly under drift, causing low discrimination and severe miscalibration (e.g., AUC around 0.518--0.666 and Brier around 0.222--0.372). The consequence is consistent with the difficulty of representing temporal non-stationarity and dynamic behavioural effects within a classical proportional hazards structure when the data-generating process evolves. XGBoost is the most competitive benchmark in our experiments. However, it still remains inferior to M1-LMISO, particularly in calibration, suggesting that model flexibility alone is insufficient when the probability scale shifts across time and regimes. Drift-oriented stream learners (HAT and ARF) generally fail to provide reliable ranking under the landmark--horizon objective, as reflected by weak AUC (often close to chance level under incremental drift) and high variance across folds, indicating a mismatch between online point-classification adaptation and survival-style risk forecasting with delayed and highly imbalanced labels.

The ablation analysis obtains three main findings. First, incorporating the longitudinal marker is the dominant source of discrimination gains, which means that the balance-deviation trajectory captures a stable risk signal that remains informative even when feature distributions change. Second, landmark encoding provides additional but moderate improvements in ranking performance (M1-LM), corresponding to its intended role as a temporal baseline adjustment that accounts for lifecycle heterogeneity and survivorship. Third, calibration-focused mechanisms are essential under drift. Both drift-aware reweighting variants (M1-TD and M1-IW) and isotonic calibration reduce probability error sharply. The quality of balancing precision and recall for the positive class is also strengthened through these techniques.

Several limitations imply opportunities for future work. Firstly, the longitudinal component in this study focuses on a single balance-based marker. A natural extension is to develop a multi-trajectory joint model that integrates additional behavioural signals to better capture heterogeneous paths to default. Secondly, while the drift settings studied here cover sudden, incremental and recurring regimes, additional realism could be achieved by incorporating macroeconomic factors such as House Price Index (HPI), mortgage rate spread and unemployment rate. Thirdly, extending the framework to competing risks (e.g., prepayment versus default) would further improve applicability in mortgage portfolios where multiple exit events co-exist. Finally, the framework can be expanded through deep architectures such as RNN-based longitudinal encoders and transformers for monthly sequences.

\newpage
\section*{Acknowledgments}
This study was supported by doctoral fellowships from China Scholarship Council (CSC) under award number 202208060251.

Authors acknowledge financial support through the project "AI4EFin AI for Energy Finance", contract number CF162/15.11.2022, financed under Romania's National Recovery and Resilience Plan, Apel nr. PNRR-III-C9-2022-I8.

ChatGPT 5.2 is used as a generative AI tool to polish the academic writing.

\section*{Declarations}
Authors have no relevant financial or non-financial interests to disclose.

This work does not contain any studies with human participants or animals performed by any of the authors. 

The raw data that support the findings of this study are available at [https://www.freddiemac.com/research/datasets/sf-loanlevel-dataset]. Processed datasets and corresponding codes can be found on GitHub under [https://doi.org/10.5281/zenodo.18392896].

The manuscript has already been uploaded to arXiv as a preprint and is available at [http://arxiv.org/abs/2601.20533].

\section*{Appendix}
\subsection*{M1-LMISO coefficients as an example}
Tables~\ref{tab:coef_variable_illustration}--\ref{tab:coef_recurring} report representative M1-LMISO coefficient summaries for the 2020 drift experiments. The coefficients refer to the logistic hazard score before isotonic calibration. Positive coefficients increase the log-odds of default within the prediction horizon, conditional on the remaining covariates, whereas negative coefficients decrease the log-odds. The large positive coefficient of $\hat{m}_i(L)$ across drift settings confirms that higher deviation from the scheduled repayment path is associated with higher future default risk.
\begin{table}[!htbp]
	\centering
	\caption{Variables in the M1-LMISO coefficient tables.}
	\label{tab:coef_variable_illustration}
	\small
	\begin{tabular}{p{3.5cm}p{10.2cm}}
		\toprule
		Variable & Interpretation \\
		\midrule
		\textit{Intercept} & Baseline log-odds term in the logistic hazard score before isotonic calibration. \\
		$\hat{m}_i(L)$ & Fitted longitudinal behavioural marker evaluated at landmark $L$, summarizing the borrower-specific deviation from the scheduled amortisation path. \\
		$L/N_i$ & Normalised loan age at the landmark, where $N_i$ denotes the original loan term of loan $i$. \\
		\textit{landmark\_k} & Landmark one-hot indicator for month $k$, capturing landmark-specific baseline risk and lifecycle heterogeneity. \\
		\textit{OrigInterestRate} & Original interest rate at loan origination. \\
		\textit{OrigLTV} & Original loan-to-value ratio. \\
		\textit{DTI} & Debt-to-income ratio at origination. \\
		\textit{OrigLoanTerm} & Original contractual loan term in months. \\
		\textit{CreditScore} & Borrower credit score at origination. \\
		\textit{NumBorrowers} & Number of borrowers obligated to repay the mortgage. \\
		\textit{Occupancy} & Occupancy one-hot indicators for investment property, primary residence and second home. \\
		\textit{LoanPurpose} & Loan-purpose one-hot indicators for cash-out refinance, no-cash-out refinance and purchase. \\
		\bottomrule
	\end{tabular}
\end{table}

\begin{table}[!htbp]
	\centering
	\caption{M1-LMISO coefficients under \textbf{sudden drift} in 2020 (mean and SD across 5-fold grouped CV).}
	\label{tab:coef_sudden}
	\small
	\begin{tabular}{lcc}
		\toprule
		Variable & Coefficient mean & Coefficient SD \\
		\midrule
		$\hat{m}_i(L)$ & 2.7866 & 0.0676 \\
		\textit{landmark\_12} & -2.3166 & 0.0326 \\
		\textit{Intercept} & 1.9568 & 0.2315 \\
		\textit{landmark\_15} & -1.8514 & 0.0279 \\
		\textit{landmark\_48} & 1.8429 & 0.0394 \\
		\textit{landmark\_45} & 1.6758 & 0.0271 \\
		\textit{landmark\_18} & -1.3973 & 0.0269 \\
		\textit{landmark\_42} & 1.3860 & 0.0255 \\
		$L/N_i$ & -1.1133 & 0.4269 \\
		\textit{landmark\_39} & 1.0648 & 0.0215 \\
		\textit{landmark\_21} & -0.9780 & 0.0183 \\
		\textit{landmark\_36} & 0.7769 & 0.0188 \\
		\textit{landmark\_24} & -0.6042 & 0.0098 \\
		\textit{landmark\_33} & 0.4643 & 0.0081 \\
		\textit{landmark\_27} & -0.2064 & 0.0062 \\
		\textit{OrigInterestRate} & -0.2024 & 0.0107 \\
		\textit{landmark\_30} & 0.1433 & 0.0096 \\
		\textit{Occupancy\_I} & 0.0771 & 0.0241 \\
		\textit{Occupancy\_P} & -0.0485 & 0.0210 \\
		\textit{NumBorrowers} & -0.0289 & 0.0209 \\
		\textit{Occupancy\_S} & -0.0287 & 0.0114 \\
		\textit{LoanPurpose\_P} & 0.0197 & 0.0137 \\
		\textit{LoanPurpose\_C} & -0.0126 & 0.0123 \\
		\textit{LoanPurpose\_N} & -0.0071 & 0.0129 \\
		\textit{OrigLTV} & -0.0020 & 0.0005 \\
		\textit{DTI} & 0.0005 & 0.0013 \\
		\textit{OrigLoanTerm} & 0.0002 & 0.0001 \\
		\textit{CreditScore} & -0.0002 & 0.0003 \\
		\bottomrule
	\end{tabular}
\end{table}

\begin{table}[!htbp]
	\centering
	\caption{M1-LMISO coefficients under \textbf{incremental drift} in 2020 (mean and SD across 5-fold grouped CV).}
	\label{tab:coef_incremental}
	\small
	\begin{tabular}{lcc}
		\toprule
		Variable & Coefficient mean & Coefficient SD \\
		\midrule
		\textit{Intercept} & 2.5448 & 0.3332 \\
		\textit{landmark\_12} & -2.4693 & 0.0189 \\
		$\hat{m}_i(L)$ & 2.1473 & 0.1259 \\
		$L/N_i$ & 2.0995 & 0.3165 \\
		\textit{landmark\_15} & -2.0654 & 0.0183 \\
		\textit{landmark\_48} & 2.0225 & 0.0393 \\
		\textit{landmark\_45} & 1.9036 & 0.0313 \\
		\textit{landmark\_18} & -1.6246 & 0.0198 \\
		\textit{landmark\_42} & 1.6056 & 0.0263 \\
		\textit{landmark\_39} & 1.2800 & 0.0114 \\
		\textit{landmark\_21} & -1.1827 & 0.0179 \\
		\textit{landmark\_36} & 0.9184 & 0.0086 \\
		\textit{landmark\_24} & -0.7531 & 0.0187 \\
		\textit{landmark\_33} & 0.5333 & 0.0111 \\
		\textit{landmark\_27} & -0.2974 & 0.0115 \\
		\textit{OrigInterestRate} & -0.1376 & 0.0244 \\
		\textit{landmark\_30} & 0.1291 & 0.0043 \\
		\textit{Occupancy\_P} & -0.1049 & 0.0224 \\
		\textit{Occupancy\_I} & 0.0595 & 0.0286 \\
		\textit{Occupancy\_S} & 0.0453 & 0.0240 \\
		\textit{LoanPurpose\_N} & 0.0287 & 0.0105 \\
		\textit{NumBorrowers} & -0.0286 & 0.0102 \\
		\textit{LoanPurpose\_C} & -0.0190 & 0.0104 \\
		\textit{LoanPurpose\_P} & -0.0097 & 0.0044 \\
		\textit{CreditScore} & -0.0020 & 0.0004 \\
		\textit{DTI} & -0.0011 & 0.0004 \\
		\textit{OrigLTV} & -0.0011 & 0.0004 \\
		\textit{OrigLoanTerm} & 0.0011 & 0.0001 \\
		\bottomrule
	\end{tabular}
\end{table}

\begin{table}[!htbp]
	\centering
	\caption{M1-LMISO coefficients under \textbf{recurring drift} in 2020 (mean and SD across 5-fold grouped CV).}
	\label{tab:coef_recurring}
	\small
	\begin{tabular}{lcc}
		\toprule
		Variable & Coefficient mean & Coefficient SD \\
		\midrule
		$\hat{m}_i(L)$ & 1.4664 & 0.0867 \\
		\textit{landmark\_12} & -1.2682 & 0.0425 \\
		\textit{Intercept} & 0.9989 & 0.3712 \\
		\textit{landmark\_48} & 0.9739 & 0.0388 \\
		\textit{landmark\_15} & -0.9688 & 0.0336 \\
		\textit{landmark\_45} & 0.9111 & 0.0323 \\
		\textit{landmark\_42} & 0.8184 & 0.0283 \\
		\textit{landmark\_18} & -0.7719 & 0.0265 \\
		\textit{landmark\_39} & 0.6603 & 0.0227 \\
		\textit{landmark\_21} & -0.6093 & 0.0196 \\
		$L/N_i$ & -0.5735 & 0.4877 \\
		\textit{landmark\_36} & 0.4208 & 0.0112 \\
		\textit{landmark\_24} & -0.3661 & 0.0130 \\
		\textit{landmark\_33} & 0.2040 & 0.0102 \\
		\textit{OrigInterestRate} & 0.1492 & 0.0388 \\
		\textit{Occupancy\_I} & -0.1207 & 0.0397 \\
		\textit{LoanPurpose\_P} & 0.1174 & 0.0278 \\
		\textit{landmark\_27} & -0.0860 & 0.0067 \\
		\textit{landmark\_30} & 0.0818 & 0.0066 \\
		\textit{NumBorrowers} & -0.0721 & 0.0117 \\
		\textit{LoanPurpose\_C} & -0.0651 & 0.0369 \\
		\textit{Occupancy\_S} & 0.0637 & 0.0341 \\
		\textit{Occupancy\_P} & 0.0570 & 0.0357 \\
		\textit{LoanPurpose\_N} & -0.0522 & 0.0206 \\
		\textit{DTI} & -0.0012 & 0.0013 \\
		\textit{CreditScore} & -0.0011 & 0.0003 \\
		\textit{OrigLoanTerm} & -0.0006 & 0.0002 \\
		\textit{OrigLTV} & 0.0004 & 0.0006 \\
		\bottomrule
	\end{tabular}
\end{table}

\clearpage
\subsection*{Comparison without drift}
Tables~\ref{tab:nodrift_2000}--\ref{tab:nodrift_2021} report the benchmark comparison when no artificial drift is imposed. In stationary periods, the default rate typically remains at an extremely low level, producing a severe class imbalance problem and uniformly poor F1 scores across models, especially in 2020 and 2021. M1-LMISO consistently improves Brier score relative to the uncalibrated M1-Joint and M1-LM variants and remains competitive in AUC, but its F1 advantage is limited under a fixed 0.5 threshold. XGBoost achieves strong F1 in some no-drift years and the highest AUC in selected cases, while its calibration is not consistently superior. These results show that the proposed framework remains well calibrated in the stationary setting and that its main empirical advantage lies in calibrated dynamic prediction under non-stationarity rather than in threshold-specific F1 optimisation under rare-event conditions.
\begin{table}[!htbp]
	\centering
	\caption{Results for year 2000 without drift (mean (SD), 5-fold grouped CV).}
	\label{tab:nodrift_2000}
	\begin{tabular}{lccc}
		\toprule
		Model & AUC & Brier & F1 \\
		\midrule
		M1 & 0.696 (0.012) & 0.221 (0.003) & 0.454 (0.010) \\
		M1-Joint & 0.731 (0.006) & 0.211 (0.003) & 0.488 (0.010) \\
		M1-LM & 0.739 (0.005) & 0.206 (0.002) & 0.493 (0.007) \\
		M1-TD & 0.727 (0.007) & 0.154 (0.004) & 0.236 (0.020) \\
		M1-IW & 0.731 (0.005) & 0.154 (0.005) & 0.233 (0.027) \\
		M1-LMISO & 0.739 (0.005) & 0.152 (0.004) & 0.314 (0.012) \\
		Cox & 0.642 (0.009) & 0.035 (0.001) & 0.000 (0.000) \\
		XGB & 0.747 (0.003) & 0.157 (0.003) & 0.469 (0.006) \\
		HAT & 0.654 (0.019) & 0.231 (0.032) & 0.455 (0.010) \\
		ARF & 0.696 (0.007) & 0.210 (0.005) & 0.454 (0.014) \\
		\bottomrule
	\end{tabular}
\end{table}

\begin{table}[!htbp]
	\centering
	\caption{Results for year 2010 without drift (mean (SD), 5-fold grouped CV).}
	\label{tab:nodrift_2010}
	\begin{tabular}{lccc}
		\toprule
		Model & AUC & Brier & F1 \\
		\midrule
		M1 & 0.732 (0.015) & 0.204 (0.003) & 0.309 (0.014) \\
		M1-Joint & 0.746 (0.011) & 0.200 (0.003) & 0.320 (0.020) \\
		M1-LM & 0.772 (0.007) & 0.191 (0.003) & 0.336 (0.017) \\
		M1-TD & 0.746 (0.011) & 0.085 (0.004) & 0.147 (0.028) \\
		M1-IW & 0.746 (0.011) & 0.085 (0.004) & 0.112 (0.019) \\
		M1-LMISO & 0.772 (0.007) & 0.083 (0.004) & 0.129 (0.021) \\
		Cox & 0.667 (0.037) & 0.017 (0.001) & 0.000 (0.000) \\
		XGB & 0.712 (0.009) & 0.092 (0.003) & 0.257 (0.028) \\
		HAT & 0.679 (0.033) & 0.114 (0.033) & 0.213 (0.034) \\
		ARF & 0.702 (0.013) & 0.096 (0.002) & 0.227 (0.017) \\
		\bottomrule
	\end{tabular}
\end{table}

\begin{table}[!htbp]
	\centering
	\caption{Results for year 2020 without drift (mean (SD), 5-fold grouped CV).}
	\label{tab:nodrift_2020}
	\begin{tabular}{lccc}
		\toprule
		Model & AUC & Brier & F1 \\
		\midrule
		M1 & 0.674 (0.008) & 0.224 (0.001) & 0.260 (0.013) \\
		M1-Joint & 0.696 (0.011) & 0.220 (0.001) & 0.272 (0.009) \\
		M1-LM & 0.696 (0.011) & 0.219 (0.001) & 0.272 (0.009) \\
		M1-TD & 0.700 (0.012) & 0.087 (0.005) & 0.016 (0.008) \\
		M1-IW & 0.699 (0.012) & 0.087 (0.005) & 0.014 (0.007) \\
		M1-LMISO & 0.696 (0.011) & 0.088 (0.005) & 0.011 (0.006) \\
		Cox & 0.648 (0.017) & 0.018 (0.001) & 0.000 (0.000) \\
		XGB & 0.703 (0.011) & 0.083 (0.004) & 0.349 (0.021) \\
		HAT & 0.671 (0.018) & 0.079 (0.005) & 0.301 (0.030) \\
		ARF & 0.681 (0.014) & 0.082 (0.004) & 0.316 (0.031) \\
		\bottomrule
	\end{tabular}
\end{table}

\begin{table}[!htbp]
	\centering
	\caption{Results for year 2021 without drift (mean (SD), 5-fold grouped CV).}
	\label{tab:nodrift_2021}
	\begin{tabular}{lccc}
		\toprule
		Model & AUC & Brier & F1 \\
		\midrule
		M1 & 0.650 (0.016) & 0.229 (0.000) & 0.201 (0.008) \\
		M1-Joint & 0.683 (0.011) & 0.222 (0.001) & 0.220 (0.010) \\
		M1-LM & 0.684 (0.011) & 0.222 (0.001) & 0.221 (0.010) \\
		M1-TD & 0.684 (0.011) & 0.070 (0.003) & 0.004 (0.005) \\
		M1-IW & 0.683 (0.011) & 0.070 (0.003) & 0.003 (0.004) \\
		M1-LMISO & 0.683 (0.011) & 0.070 (0.003) & 0.004 (0.006) \\
		Cox & 0.634 (0.022) & 0.020 (0.001) & 0.000 (0.000) \\
		XGB & 0.649 (0.007) & 0.075 (0.004) & 0.195 (0.011) \\
		HAT & 0.655 (0.007) & 0.069 (0.004) & 0.118 (0.030) \\
		ARF & 0.634 (0.013) & 0.072 (0.003) & 0.094 (0.021) \\
		\bottomrule
	\end{tabular}
\end{table}

\clearpage
\subsection*{Sensitivity analysis using \textit{CurLoanDel} $\geq 3$}
Tables~\ref{tab:ge3_sudden}--\ref{tab:ge3_recurring} report the 2020 sensitivity analysis under the common 90-days-plus delinquency definition. The hazard-based models retain moderate discrimination, but F1 drops sharply under the fixed 0.5 classification threshold because severe delinquency events are sparse, making the comparison among F1 scores meaningless. This indicates that the threshold-specific model tuning and validation are required in credit scoring because the choice of default setting does change the prediction task and performance.
\begin{table}[!htbp]
	\centering
	\caption{Results for year 2020 under \textbf{sudden drift} using \textit{CurLoanDel} $\geq 3$ (mean (SD), 5-fold grouped CV).}
	\label{tab:ge3_sudden}
	\begin{tabular}{lccc}
		\toprule
		Model & AUC & Brier & F1 \\
		\midrule
		M1 & 0.785 (0.013) & 0.189 (0.001) & 0.134 (0.011) \\
		M1-Joint & 0.795 (0.013) & 0.185 (0.002) & 0.136 (0.015) \\
		M1-LM & 0.795 (0.013) & 0.185 (0.002) & 0.137 (0.015) \\
		M1-TD & 0.800 (0.012) & 0.027 (0.002) & 0.020 (0.012) \\
		M1-IW & 0.797 (0.013) & 0.028 (0.002) & 0.008 (0.006) \\
		M1-LMISO & 0.793 (0.013) & 0.028 (0.002) & 0.007 (0.006) \\
		Cox & 0.753 (0.016) & 0.003 (0.001) & 0.000 (0.000) \\
		XGB & 0.888 (0.008) & 0.018 (0.001) & 0.603 (0.025) \\
		HAT & 0.795 (0.025) & 0.020 (0.002) & 0.523 (0.036) \\
		ARF & 0.843 (0.014) & 0.022 (0.002) & 0.243 (0.103) \\
		\bottomrule
	\end{tabular}
\end{table}

\begin{table}[!htbp]
	\centering
	\caption{Results for year 2020 under \textbf{incremental drift} using \textit{CurLoanDel} $\geq 3$ (mean (SD), 5-fold grouped CV).}
	\label{tab:ge3_incremental}
	\begin{tabular}{lccc}
		\toprule
		Model & AUC & Brier & F1 \\
		\midrule
		M1 & 0.788 (0.012) & 0.187 (0.001) & 0.135 (0.011) \\
		M1-Joint & 0.820 (0.015) & 0.171 (0.001) & 0.151 (0.014) \\
		M1-LM & 0.821 (0.015) & 0.170 (0.001) & 0.151 (0.014) \\
		M1-TD & 0.828 (0.013) & 0.025 (0.002) & 0.084 (0.038) \\
		M1-IW & 0.823 (0.014) & 0.027 (0.002) & 0.026 (0.015) \\
		M1-LMISO & 0.820 (0.015) & 0.027 (0.002) & 0.040 (0.024) \\
		Cox & 0.754 (0.016) & 0.003 (0.001) & 0.000 (0.000) \\
		XGB & 0.871 (0.009) & 0.018 (0.002) & 0.578 (0.018) \\
		HAT & 0.707 (0.020) & 0.264 (0.087) & 0.121 (0.021) \\
		ARF & 0.690 (0.018) & 0.278 (0.069) & 0.091 (0.016) \\
		\bottomrule
	\end{tabular}
\end{table}

\begin{table}[!htbp]
	\centering
	\caption{Results for year 2020 under \textbf{recurring drift} using \textit{CurLoanDel} $\geq 3$ (mean (SD), 5-fold grouped CV).}
	\label{tab:ge3_recurring}
	\begin{tabular}{lccc}
		\toprule
		Model & AUC & Brier & F1 \\
		\midrule
		M1 & 0.782 (0.011) & 0.190 (0.001) & 0.132 (0.009) \\
		M1-Joint & 0.805 (0.013) & 0.180 (0.002) & 0.142 (0.014) \\
		M1-LM & 0.805 (0.013) & 0.180 (0.002) & 0.142 (0.014) \\
		M1-TD & 0.815 (0.014) & 0.026 (0.002) & 0.033 (0.017) \\
		M1-IW & 0.808 (0.014) & 0.027 (0.003) & 0.010 (0.009) \\
		M1-LMISO & 0.804 (0.013) & 0.028 (0.002) & 0.009 (0.007) \\
		Cox & 0.749 (0.021) & 0.003 (0.001) & 0.000 (0.000) \\
		XGB & 0.898 (0.007) & 0.016 (0.001) & 0.660 (0.014) \\
		HAT & 0.820 (0.020) & 0.017 (0.002) & 0.629 (0.031) \\
		ARF & 0.866 (0.018) & 0.018 (0.002) & 0.547 (0.071) \\
		\bottomrule
	\end{tabular}
\end{table}

\clearpage
\subsection*{Results for other years under drifts}
\begin{table}[!htbp]
	\centering
	\caption{Results for year 2000 under sudden drift (mean (SD), 5-fold grouped CV).}
	\label{tab:2000_sudden}
	\begin{tabular}{lccc}
		\toprule
		Model & AUC & Brier & F1 \\
		\midrule
		M1 & 0.681 (0.012) & 0.226 (0.004) & 0.649 (0.010) \\
		M1-Joint & 0.740 (0.005) & 0.211 (0.002) & 0.654 (0.006) \\
		M1-LM & 0.748 (0.005) & 0.204 (0.002) & 0.685 (0.005) \\
		M1-TD & 0.732 (0.005) & 0.213 (0.001) & 0.690 (0.006) \\
		M1-IW & 0.741 (0.005) & 0.210 (0.002) & 0.675 (0.006) \\
		M1-LMISO & 0.748 (0.005) & 0.203 (0.002) & 0.695 (0.005) \\
		Cox & 0.663 (0.004) & 0.108 (0.001) & 0.000 (0.000) \\
		XGB & 0.750 (0.007) & 0.204 (0.003) & 0.689 (0.005) \\
		HAT & 0.576 (0.028) & 0.339 (0.028) & 0.683 (0.006) \\
		ARF & 0.564 (0.010) & 0.358 (0.010) & 0.684 (0.006) \\
		\bottomrule
	\end{tabular}
\end{table}

\begin{table}[!htbp]
	\centering
	\caption{Results for year 2000 under incremental drift (mean (SD), 5-fold grouped CV).}
	\label{tab:2000_incremental}
	\begin{tabular}{lccc}
		\toprule
		Model & AUC & Brier & F1 \\
		\midrule
		M1 & 0.700 (0.006) & 0.216 (0.002) & 0.588 (0.010) \\
		M1-Joint & 0.746 (0.004) & 0.199 (0.001) & 0.622 (0.005) \\
		M1-LM & 0.749 (0.004) & 0.195 (0.002) & 0.612 (0.005) \\
		M1-TD & 0.744 (0.004) & 0.191 (0.001) & 0.583 (0.005) \\
		M1-IW & 0.746 (0.004) & 0.191 (0.001) & 0.572 (0.004) \\
		M1-LMISO & 0.749 (0.004) & 0.188 (0.002) & 0.580 (0.006) \\
		Cox & 0.707 (0.012) & 0.052 (0.001) & 0.000 (0.000) \\
		XGB & 0.749 (0.005) & 0.192 (0.001) & 0.617 (0.007) \\
		HAT & 0.539 (0.026) & 0.510 (0.013) & 0.569 (0.006) \\
		ARF & 0.551 (0.017) & 0.501 (0.006) & 0.569 (0.006) \\
		\bottomrule
	\end{tabular}
\end{table}

\begin{table}[!htbp]
	\centering
	\caption{Results for year 2000 under recurring drift (mean (SD), 5-fold grouped CV).}
	\label{tab:2000_recurring}
	\begin{tabular}{lccc}
		\toprule
		Model & AUC & Brier & F1 \\
		\midrule
		M1 & 0.577 (0.015) & 0.243 (0.002) & 0.657 (0.010) \\
		M1-Joint & 0.616 (0.013) & 0.239 (0.002) & 0.664 (0.006) \\
		M1-LM & 0.627 (0.012) & 0.237 (0.001) & 0.669 (0.006) \\
		M1-TD & 0.610 (0.012) & 0.196 (0.003) & 0.835 (0.005) \\
		M1-IW & 0.615 (0.013) & 0.196 (0.003) & 0.835 (0.005) \\
		M1-LMISO & 0.627 (0.012) & 0.194 (0.003) & 0.835 (0.005) \\
		Cox & 0.659 (0.006) & 0.075 (0.000) & 0.000 (0.000) \\
		XGB & 0.615 (0.011) & 0.219 (0.004) & 0.737 (0.007) \\
		HAT & 0.548 (0.020) & 0.215 (0.009) & 0.833 (0.008) \\
		ARF & 0.554 (0.016) & 0.212 (0.005) & 0.833 (0.006) \\
		\bottomrule
	\end{tabular}
\end{table}

\begin{table}[!htbp]
	\centering
	\caption{Results for year 2010 under sudden drift (mean (SD), 5-fold grouped CV).}
	\label{tab:2010_sudden}
	\begin{tabular}{lccc}
		\toprule
		Model & AUC & Brier & F1 \\
		\midrule
		M1 & 0.799 (0.005) & 0.184 (0.003) & 0.825 (0.004) \\
		M1-Joint & 0.871 (0.003) & 0.157 (0.002) & 0.829 (0.002) \\
		M1-LM & 0.874 (0.003) & 0.154 (0.001) & 0.838 (0.002) \\
		M1-TD & 0.871 (0.003) & 0.107 (0.003) & 0.903 (0.003) \\
		M1-IW & 0.871 (0.004) & 0.107 (0.003) & 0.903 (0.003) \\
		M1-LMISO & 0.874 (0.003) & 0.105 (0.003) & 0.905 (0.003) \\
		Cox & 0.685 (0.005) & 0.251 (0.002) & 0.189 (0.007) \\
		XGB & 0.863 (0.003) & 0.128 (0.002) & 0.878 (0.002) \\
		HAT & 0.701 (0.082) & 0.160 (0.014) & 0.898 (0.004) \\
		ARF & 0.497 (0.030) & 0.182 (0.006) & 0.899 (0.003) \\
		\bottomrule
	\end{tabular}
\end{table}

\begin{table}[!htbp]
	\centering
	\caption{Results for year 2010 under incremental drift (mean (SD), 5-fold grouped CV).}
	\label{tab:2010_incremental}
	\begin{tabular}{lccc}
		\toprule
		Model & AUC & Brier & F1 \\
		\midrule
		M1 & 0.805 (0.001) & 0.182 (0.002) & 0.787 (0.002) \\
		M1-Joint & 0.842 (0.002) & 0.167 (0.002) & 0.808 (0.002) \\
		M1-LM & 0.847 (0.002) & 0.167 (0.002) & 0.795 (0.002) \\
		M1-TD & 0.842 (0.002) & 0.138 (0.002) & 0.856 (0.002) \\
		M1-IW & 0.842 (0.002) & 0.138 (0.001) & 0.857 (0.002) \\
		M1-LMISO & 0.847 (0.002) & 0.135 (0.001) & 0.854 (0.001) \\
		Cox & 0.681 (0.006) & 0.176 (0.001) & 0.142 (0.005) \\
		XGB & 0.834 (0.003) & 0.156 (0.002) & 0.821 (0.002) \\
		HAT & 0.526 (0.052) & 0.247 (0.006) & 0.853 (0.003) \\
		ARF & 0.500 (0.030) & 0.255 (0.005) & 0.854 (0.003) \\
		\bottomrule
	\end{tabular}
\end{table}

\begin{table}[!htbp]
	\centering
	\caption{Results for year 2010 under recurring drift (mean (SD), 5-fold grouped CV).}
	\label{tab:2010_recurring}
	\begin{tabular}{lccc}
		\toprule
		Model & AUC & Brier & F1 \\
		\midrule
		M1 & 0.709 (0.024) & 0.219 (0.003) & 0.742 (0.006) \\
		M1-Joint & 0.748 (0.009) & 0.212 (0.001) & 0.737 (0.002) \\
		M1-LM & 0.753 (0.009) & 0.210 (0.001) & 0.744 (0.002) \\
		M1-TD & 0.747 (0.009) & 0.078 (0.002) & 0.952 (0.001) \\
		M1-IW & 0.749 (0.010) & 0.078 (0.002) & 0.952 (0.001) \\
		M1-LMISO & 0.752 (0.009) & 0.078 (0.002) & 0.952 (0.002) \\
		Cox & 0.541 (0.007) & 0.177 (0.002) & 0.000 (0.000) \\
		XGB & 0.722 (0.010) & 0.106 (0.003) & 0.918 (0.002) \\
		HAT & 0.584 (0.059) & 0.095 (0.011) & 0.946 (0.006) \\
		ARF & 0.524 (0.020) & 0.090 (0.003) & 0.952 (0.002) \\
		\bottomrule
	\end{tabular}
\end{table}

\begin{table}[!htbp]
	\centering
	\caption{Results for year 2021 under sudden drift (mean (SD), 5-fold grouped CV).}
	\label{tab:2021_sudden}
	\begin{tabular}{lccc}
		\toprule
		Model & AUC & Brier & F1 \\
		\midrule
		M1 & 0.650 (0.010) & 0.230 (0.001) & 0.813 (0.015) \\
		M1-Joint & 0.712 (0.004) & 0.219 (0.001) & 0.753 (0.001) \\
		M1-LM & 0.717 (0.004) & 0.216 (0.001) & 0.763 (0.002) \\
		M1-TD & 0.712 (0.004) & 0.087 (0.003) & 0.946 (0.002) \\
		M1-IW & 0.712 (0.004) & 0.088 (0.003) & 0.946 (0.002) \\
		M1-LMISO & 0.717 (0.004) & 0.087 (0.003) & 0.946 (0.002) \\
		Cox & 0.517 (0.003) & 0.383 (0.001) & 0.344 (0.002) \\
		XGB & 0.673 (0.004) & 0.114 (0.002) & 0.915 (0.002) \\
		HAT & 0.569 (0.047) & 0.095 (0.004) & 0.945 (0.002) \\
		ARF & 0.512 (0.008) & 0.096 (0.003) & 0.946 (0.002) \\
		\bottomrule
	\end{tabular}
\end{table}

\begin{table}[!htbp]
	\centering
	\caption{Results for year 2021 under incremental drift (mean (SD), 5-fold grouped CV).}
	\label{tab:2021_incremental}
	\begin{tabular}{lccc}
		\toprule
		Model & AUC & Brier & F1 \\
		\midrule
		M1 & 0.749 (0.002) & 0.207 (0.000) & 0.766 (0.001) \\
		M1-Joint & 0.750 (0.002) & 0.207 (0.000) & 0.768 (0.001) \\
		M1-LM & 0.756 (0.002) & 0.203 (0.000) & 0.785 (0.002) \\
		M1-TD & 0.750 (0.002) & 0.120 (0.002) & 0.912 (0.002) \\
		M1-IW & 0.749 (0.002) & 0.121 (0.002) & 0.913 (0.002) \\
		M1-LMISO & 0.756 (0.002) & 0.119 (0.002) & 0.913 (0.002) \\
		Cox & 0.609 (0.003) & 0.378 (0.001) & 0.383 (0.001) \\
		XGB & 0.720 (0.003) & 0.150 (0.002) & 0.866 (0.003) \\
		HAT & 0.411 (0.052) & 0.155 (0.003) & 0.913 (0.002) \\
		ARF & 0.515 (0.011) & 0.146 (0.002) & 0.913 (0.002) \\
		\bottomrule
	\end{tabular}
\end{table}

\begin{table}[!htbp]
	\centering
	\caption{Results for year 2021 under recurring drift (mean (SD), 5-fold grouped CV).}
	\label{tab:2021_recurring}
	\begin{tabular}{lccc}
		\toprule
		Model & AUC & Brier & F1 \\
		\midrule
		M1 & 0.577 (0.004) & 0.246 (0.001) & 0.682 (0.006) \\
		M1-Joint & 0.626 (0.005) & 0.239 (0.000) & 0.703 (0.001) \\
		M1-LM & 0.629 (0.004) & 0.237 (0.000) & 0.701 (0.002) \\
		M1-TD & 0.625 (0.005) & 0.113 (0.003) & 0.929 (0.002) \\
		M1-IW & 0.625 (0.004) & 0.113 (0.003) & 0.929 (0.002) \\
		M1-LMISO & 0.628 (0.004) & 0.113 (0.003) & 0.929 (0.002) \\
		Cox & 0.512 (0.005) & 0.227 (0.001) & 0.012 (0.022) \\
		XGB & 0.576 (0.003) & 0.142 (0.003) & 0.896 (0.003) \\
		HAT & 0.524 (0.018) & 0.122 (0.006) & 0.925 (0.006) \\
		ARF & 0.506 (0.008) & 0.119 (0.004) & 0.929 (0.003) \\
		\bottomrule
	\end{tabular}
\end{table}

\clearpage
\bibliographystyle{plainnat}
\bibliography{references}
\end{document}